\documentclass[conference]{IEEEtran}
\IEEEoverridecommandlockouts
\usepackage{cite}
\usepackage{amsmath,amssymb,amsfonts}
\usepackage{algorithmic}
\usepackage{graphicx}
\usepackage{textcomp}
\usepackage{hyperref}
\usepackage{subfig}
\usepackage{multirow}
\usepackage{xcolor}
\def\BibTeX{{\rm B\kern-.05em{\sc i\kern-.025em b}\kern-.08em
    T\kern-.1667em\lower.7ex\hbox{E}\kern-.125emX}}
\begin{document}

\title{Neurosymbolic AI for Travel Demand Prediction: Integrating Decision Tree Rules into Neural Networks
\thanks{This material is based upon work supported by the NASA Aeronautics Research Mission Directorate (ARMD) University Leadership Initiative (ULI) under cooperative agreement number 80NSSC23M0059. This research was also partially supported by the U.S. National Science Foundation through Grant No. 2317117 and Grant No. 2309760.}
}

\author{\IEEEauthorblockN{1\textsuperscript{st} Kamal Acharya}
\IEEEauthorblockA{\textit{Department of Information Systems} \\
\textit{University of Maryland Baltimore County}\\
Baltimore, MD, US \\
kamala2@umbc.edu}
\and
\IEEEauthorblockN{2\textsuperscript{nd} Mehul Lad}
\IEEEauthorblockA{\textit{Department of Information Systems} \\
\textit{University of Maryland Baltimore County}\\
Baltimore, MD, US \\
du72811@umbc.edu}
\and
\IEEEauthorblockN{3\textsuperscript{rd} Liang Sun}
\IEEEauthorblockA{\textit{Department of Mechanical Engineering} \\
\textit{Baylor University}\\
Waco,TX, US \\
liang\_sun@baylor.edu}
\and
\IEEEauthorblockN{4\textsuperscript{th} Houbing Song}
\IEEEauthorblockA{\textit{Department of Information Systems} \\
\textit{University of Maryland Baltimore County}\\
Baltimore, MD, US \\
songh@umbc.edu}

}

\maketitle

\begin{abstract}
Travel demand prediction is crucial for optimizing transportation planning, resource allocation, and infrastructure development, ensuring efficient mobility and economic sustainability. This study introduces a Neurosymbolic Artificial Intelligence (Neurosymbolic AI) framework that integrates decision tree (DT)-based symbolic rules with neural networks (NNs) to predict travel demand, leveraging the interpretability of symbolic reasoning and the predictive power of neural learning. The framework utilizes data from diverse sources, including geospatial, economic, and mobility datasets, to build a comprehensive feature set. DTs are employed to extract interpretable if-then rules that capture key patterns, which are then incorporated as additional features into a NN to enhance its predictive capabilities. Experimental results show that the combined dataset, enriched with symbolic rules, consistently outperforms standalone datasets across multiple evaluation metrics, including Mean Absolute Error (MAE), \(R^2\), and Common Part of Commuters (CPC). Rules selected at finer variance thresholds (e.g., 0.0001) demonstrate superior effectiveness in capturing nuanced relationships, reducing prediction errors, and aligning with observed commuter patterns. By merging symbolic and neural learning paradigms, this Neurosymbolic approach achieves both interpretability and accuracy. The data and code can be accessed on GitHub. \footnote{Github: \url{https://github.com/lotussavy/ICNS-2025.git }}
\end{abstract}

\begin{IEEEkeywords}
Decision Tree, Neural Network, Neurosymbolic AI, Travel Demand Prediction
\end{IEEEkeywords}

\section{Introduction}
\label{sec:introduction}
Travel demand prediction plays a critical role in transportation planning, infrastructure development, and resource allocation\cite{banerjee2020passenger}. Accurate forecasting of travel demand between regions, such as counties, enables stakeholders to optimize transportation networks, reduce congestion, and make informed decisions regarding public infrastructure investments. As urbanization accelerates and mobility patterns become increasingly complex, robust and reliable methods for demand prediction are essential for meeting the dynamic needs of commuters while improving the efficiency of transportation systems.

Predicting travel demand between counties presents a significant challenge, as it requires modeling complex relationships between spatial, economic, and mobility factors. Traditional methods, such as statistical and time-series models\cite{saadallah2018bright}, often struggle to capture the nonlinear and multidimensional interactions inherent in real-world travel data\cite{8489530}. While these approaches provide simplicity and interpretability, they fail to address the intricate dependencies that drive travel demand. Neural networks (NNs), on the other hand, excel at learning these complex patterns due to their advanced capabilities. However, their "black-box" nature limits interpretability, which is critical in transportation applications where transparency and accountability are essential. With the increasing availability of high-resolution mobility and economic datasets\cite{adrian_tantau__2020}\cite{cheng2020integrating}, there is a growing demand for advanced methodologies that can effectively uncover and model these hidden patterns while maintaining a balance between accuracy and interpretability.

\begin{figure}[htbp]
\centering
\includegraphics[width=0.5\textwidth]{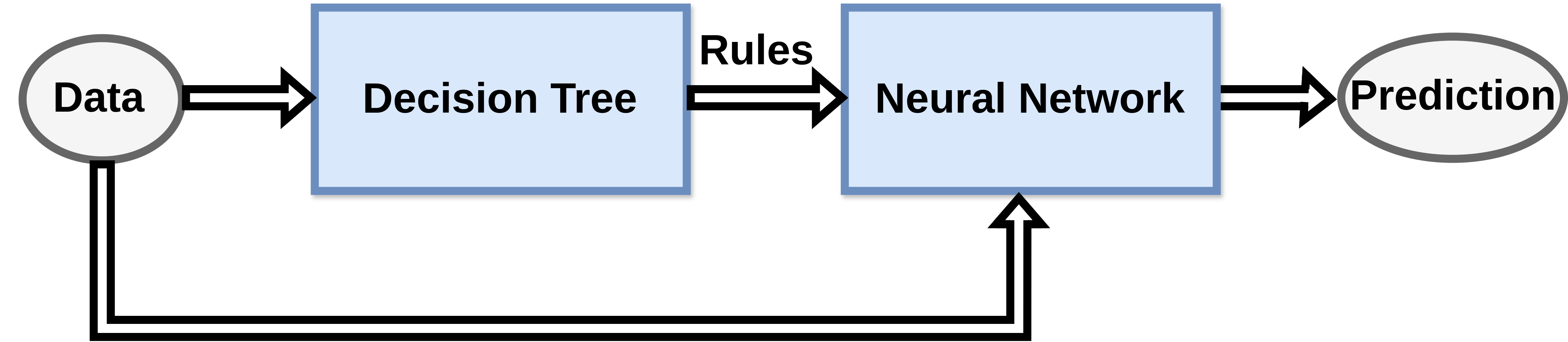}
\caption{Neurosymbolic Approach Used in the Research}
\label{fig:NeurosymbolicModel}
\end{figure}

In this paper, we propose a novel Neurosymbolic approach as shown in the \autoref{fig:NeurosymbolicModel} that integrates decision tree (DT) rules with NNs to enhance both the interpretability and predictive power of travel demand prediction models. DTs are used to extract interpretable if-then rules that capture key travel patterns and relationships between features, while NNs are employed to model nonlinear interactions and improve accuracy. By encoding DT rules as additional input features for the NN, we bridge the gap between symbolic and neural learning paradigms. Extensive experiments demonstrate that our approach significantly improves predictive performance, as measured by metrics such as Mean Absolute Error (MAE), \(R^2\), and Common Part of Commuters (CPC), especially when using rules selected with fine variance thresholds.

Our work represents a significant advancement in demand modeling by integrating the interpretability of DTs with the predictive accuracy of NNs within a unified framework. While previous research has largely focused on neural approaches or, in some cases, independently explored symbolic methods, limited attention has been given to explainable artificial intelligence (XAI) \cite{arrieta2020explainable} in this context. Crucially, no prior studies have investigated the application of Neurosymbolic Artificial Intelligence (Neurosymbolic AI) \cite{garcez2023neurosymbolic} in the domain of travel demand modeling. This study is the first to explore the synergy between NNs and symbolic models for predicting travel demand. Further, our methodology incorporates variance-based rule selection, providing a novel mechanism to effectively balance model complexity with interpretability, setting it apart from existing hybrid approaches.

The remainder of this paper is structured as follows: Section~\ref{sec:related_work} reviews related work on travel demand prediction. Section~\ref{sec:methodology} details the proposed methodology, including data preprocessing, rule extraction, and NN integration. Section~\ref{sec:experiments} describes the evaluation metrics and presents results. Section~\ref{sec:discussion} discusses the findings and their implications. Finally, Section~\ref{sec:conclusion} concludes the paper and outlines directions for future research.

\section{Related Work}
\label{sec:related_work}

Traditional methods for predicting travel demand have relied heavily on statistical models, such as gravity models\cite{jung2008gravity} and logistic regression techniques\cite{wei2015logistic}. Regression analysis has also been utilized to predict the demand for Regional Air Mobility (RAM)\cite{acharya2025regional} and Advanced Air Mobility (AAM)\cite{10825121}. These models often assume linear relationships and may not effectively capture the complex, nonlinear interactions present in transportation systems.

With the advent of machine learning, more sophisticated models have been employed to improve prediction accuracy. NNs have been widely used due to their ability to model nonlinear relationships and learn from large datasets \cite{guo2020residual}. In the context of travel demand prediction, NNs have demonstrated superior performance over traditional statistical methods by capturing intricate patterns in mobility data \cite{rajendran2021predicting}. They have also been employed as a supportive tool to enhance genetic algorithms, leading to improved outcomes\cite{acharya2024improving}. However, a significant limitation of NNs is their lack of interpretability. The "black-box" nature of these models makes it challenging for practitioners to understand the underlying decision-making processes, which is crucial in transportation planning where transparency and explainability are essential\cite{golshani2018modeling}. To address this black-box puzzle, two distinct branches of AI come into play: XAI\cite{arrieta2020explainable} and Neurosymbolic AI\cite{garcez2023neurosymbolic}. XAI focuses on providing explanations for AI models, typically after the training process. In contrast, Neurosymbolic AI integrates neural learning with symbolic reasoning directly within the model's architecture, creating a more inherently interpretable framework.

Various recent research has explored the application of XAI in various aspects of trip demand prediction, from forecasting overall travel demand to understanding individual mode choices. Hu et al.\cite{hu2023examining} tackled the challenge of predicting population inflow using mobile device location data, comparing various tree-based models and interpretation techniques. Their study revealed that boosting trees outperformed other models and that while feature importance rankings were consistent across models, the choice of importance measures and hyperparameter settings did influence the results. Moving from a nationwide perspective to a city-wide focus, Kim et al. \cite{kim2020stepwise} investigated the dynamic relationship between taxi and ride-hailing services in New York City. Their two-stage modeling approach combined linear regression with a long short-term memory network to predict taxi demand, effectively capturing the influence of ride-hailing services, day of the week, weather conditions, and holidays. This study demonstrated the potential of interpretable deep learning models for developing active demand management strategies, such as quota control systems, to balance demand between different modes and mitigate congestion. Kim\cite{kim2021analysis} examined travel mode choice in Seoul, employing extreme gradient boosting (XGB) \cite{chen2015xgboost} and interpretation techniques like variable importance, interaction analysis, and accumulated local eﬀects (ALE) \cite{danesh2022interpretability} plots. The study not only demonstrated the accuracy of XGB but also revealed the importance of trip- and tour-related variables, age, and number of trips on tour in influencing mode choice decisions. Another study\cite{kashifi2022predicting}, achieves explainable or interpretable AI by employing two key techniques: variable importance analysis and SHapley Additive exPlanations (SHAP) \cite{lundberg2017unified} analysis. Variable importance analysis identifies the input variables, such as trip distance, traveler’s age, and weather conditions, that have the most significant influence on the model's predictions, revealing the strongest predictors of travel mode choice. SHAP analysis then quantifies the impact of each feature, like annual income and car/bicycle ownership, on the prediction for individual instances, providing a deeper understanding of how these factors work together to shape travel mode decisions.




In the transportation domain, no studies have been found that focused on Neurosymbolic approaches for travel demand prediction. Despite the potential advantages, there is a gap in the literature regarding the application of Neurosymbolic AI to travel demand prediction. Existing studies have only investigated post training accuracy and interpretability of demand prediction models. Our work addresses this gap by proposing a Neurosymbolic framework that leverages DT-extracted rules alongside NNs to predict travel demand between counties.

\section{Methodology}
\label{sec:methodology}
\begin{figure*}[htbp]
\centering
\includegraphics[width=0.9\textwidth]{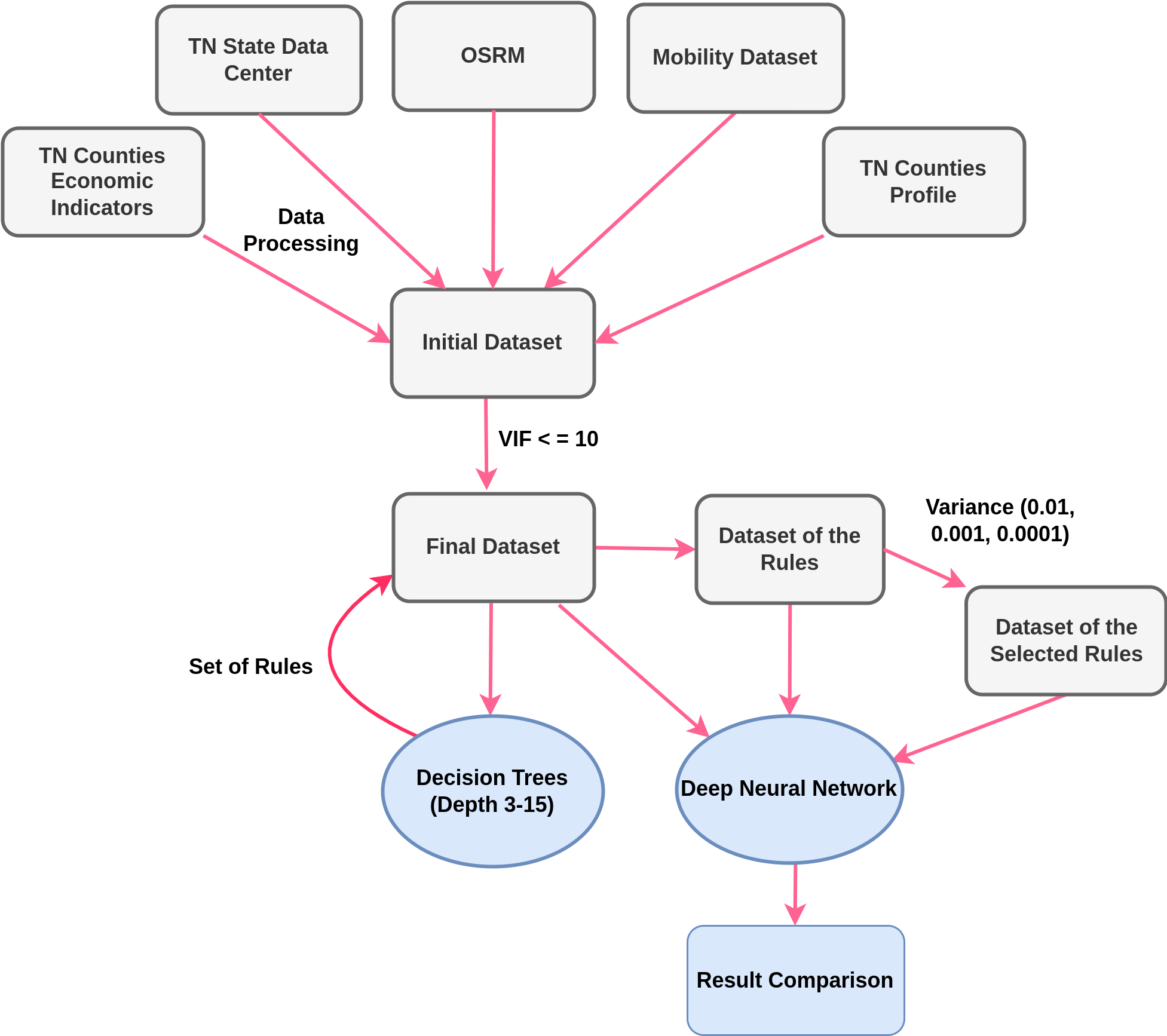}
\caption{Methodology of the Research}
\label{fig:Methodology}
\end{figure*}

The \autoref{fig:Methodology} illustrates the methodological workflow for predicting travel demand using DT and NN. It begins with data integration from diverse sources, including geospatial, economic, and mobility datasets, to form the Initial Dataset. This dataset is processed and refined into the Final Dataset, ensuring feature relevance and quality. Rules are extracted using DTs with varying depths, resulting in a Dataset of the Rules, which is further refined based on variance thresholds to produce the Dataset of the Selected Rules.

\subsection{Data Collection and Preprocessing}
The data for this study was collected from various sources to ensure the inclusion of key factors influencing trip distribution. \autoref{tab:Datasets} lists the various datasets used in this research. 

\begin{table*}
\caption{ Datasets Used in the Research}
\label{tab:Datasets}
\centering
\begin{tabular}{|c|c|}
\hline
\textbf{Name }& \textbf{Details}\\ \hline

Mobility Dataset & Multiscale Dynamic Human Mobility Flow Dataset\cite{kang2020multiscale}\\ \hline

 Population Dataset & Tennessee State Data Center\cite{TennesseeStateDataCenter}. \\ \hline

 Distance and Time Between Counties & Open Source Routing Machine (OSRM)\cite{luxen-vetter-2011}\\
 \hline

 Geographical features & Open Source Routing Machine (OSRM)\cite{luxen-vetter-2011}\\
\hline

 Economic features & Tennessee Economic Indicators\cite{tennessee_economic_2021}\\
\hline

 Tennessee County Profiles & Tennessee Commission on Children and Youth\cite{county_profiles_child_2021}\\
\hline

\end{tabular}
\end{table*}

The multiscale dynamic human mobility flow dataset \cite{kang2020multiscale} provides extensive spatiotemporal data on U.S. population movements since January 1, 2019. Compiled from anonymized mobile phone location data sourced by SafeGraph, this dataset documents origin-to-destination (O-D) population flows at the census tract level on daily bases. Its detailed granularity and temporal coverage make it a valuable resource for applications in transportation planning. For this study, the daily dataset from March 15, 2021, to April 15, 2021, was extracted. The Open Source Routing Machine (OSRM) was utilized to obtain geographical features for counties within Tennessee, while economic data was sourced from the Boyd Center for Business and Economic Research. County profiles and rankings were obtained from datasets provided by the Tennessee Commission on Children and Youth. All datasets underwent cleaning, standardization, and were merged using county-level identifiers. Missing values were imputed using median substitution, and numerical variables were scaled to facilitate comparability across features. Ultimately, 11 features were prepared for each county.

\begin{itemize}
    \item \textbf{Land Use Counts:} Consists of two features: NaturalAreaCounts and PublicAreaCounts representing natural area counts like  forest, agricultural etc. and public area counts which includes places like residential, commercial, industrial etc. respectively.
    \item \textbf{Points of Interest (POI):} Counts of POIs like educational places, commercial, healthcare, entertainment etc. within the counties.
    \item \textbf{Roads:} Classified into two categories: MajorRoads and OtherRoads.
    \item \textbf{Transportation:} Includes the total count of airports, railway stations, bus stations, etc.
    \item \textbf{Economic Features:} Consists of three features: Unemployment Rate, Employed Population, and Sales Tax Revenue.
    \item \textbf{Ranking:} Overall ranking of the counties in TN based on health, economic well-being, education etc.
    \item \textbf{Population:} Provides population count for each county.
\end{itemize}
Above features of both origin and destination counties were merged with the mobility dataset to obtain the initial dataset which consists of 22 columns for representing the features of origin and destination, 2 columns that provides time and distance required to travel from one county to another, one column that specify the day either weekday or weekend and last column provide the number of flows between the counties.

A multicollinearity check was performed to ensure that all features in the final dataset had a variance inflation factor (VIF) $\leq$ 10. This step eliminated redundant or highly correlated variables. After doing so we end up with the dataset that contains 10 features along with the population flow column, making it an 11 column dataset.

\subsection{Decision Tree Rule Extraction}
We employed DT as a method to derive interpretable rules that elucidate the relationships between features and target variables. The process of rule extraction and subsequent feature generation involves several stages, as outlined below:

\subsubsection{Construction of Decision Trees with Various Depths}
DTs of varying depths, ranging from 3 to 15, were constructed to balance the trade-off between model interpretability and predictive accuracy. Shallow trees (e.g., depth 3) emphasize interpretability by isolating the most significant features driving the outcomes, while deeper trees (e.g., depth 15) capture complex, nonlinear relationships. This variation enabled a comparative evaluation of model complexity versus performance, ensuring the retention of valuable information from both simple and complex feature interactions.

\subsubsection{Extraction of Rules from Decision Trees}
After training the DTs, the paths from root nodes to leaf nodes were traversed to extract if-then rules. Each path represents a rule delineating a subset of the data based on conditions imposed on input features. These rules encapsulate the underlying patterns and thresholds, enabling a clear understanding of the decision-making process for specific predictions. \autoref{tab:tree_rules} shows the rules that were extracted from the DT of depth 3.

\renewcommand{\arraystretch}{1.25} 
\begin{table}[htbp]
\caption{8 Rules Extracted from Decision Tree of Depth 3}
\begin{center}
\label{tab:tree_rules}
\begin{tabular}{|p{6.5cm}|c|}
\hline
\textbf{Conditions}       & \textbf{Mean Value} \\ \hline
 distance\_miles $\leq$ 46.08 AND POIs\_Destination $>$ 323.0 AND POIs\_Origin $>$ 307.0               & 31,867.81                \\ \hline
 distance\_miles $\leq$ 46.08 AND POIs\_Destination $>$ 323.0 AND POIs\_Origin $\leq$ 307.0            & 9,709.81                 \\ \hline
 distance\_miles $\leq$ 46.08 AND 323.0 $\geq$ POIs\_Destination $>$ 243.0                             & 6,492.03                 \\ \hline
 46.08 $<$ distance\_miles $\leq$ 58.77 AND NaturalAreaCounts\_Destination $>$ 935.5                  & 2,728.70                 \\ \hline
 distance\_miles $\leq$ 46.08 AND POIs\_Destination $\leq$ 243.0                                      & 1,648.22                 \\ \hline
 distance\_miles $>$ 58.77 AND POIs\_Destination $>$ 773.5                                            & 544.09                   \\ \hline
 46.08 $<$ distance\_miles $\leq$ 58.77 AND NaturalAreaCounts\_Destination $\leq$ 935.5               & 515.81                   \\ \hline
distance\_miles $>$ 58.77 AND POIs\_Destination $\leq$ 773.5                                         & 106.32                   \\ \hline
\end{tabular}
\end{center}
\end{table}

\subsection{Generating Dataset of the Rules}

The extracted rules from the DT were then applied to the dataset to generate a new set of features. These features were encoded as binary indicators, where each feature corresponds to a rule: a value of 1 indicates that the rule's conditions are met, and a value of 0 indicates otherwise. We generated various datasets corresponding to the various depths and different number of rules from it. 


To ensure the relevance and practicality of the newly generated rule-based features, a systematic filtering process was implemented. Rules with minimal variance—specifically those falling below thresholds of 0.01, 0.001, and 0.0001—were excluded. As a result, we have four different datasets for rules: one that has all rules and other three with the rules having variances of 0.01, 0.001 and 0.0001 respectively for different depth of DTs.

The resulting dataset of rules served as an intermediate output, representing a refined and interpretable feature set. This dataset was subsequently integrated into a NN model for further predictive analysis. 

\subsection{Neural Network Implementation}

To integrate the rules into the NN, we generated 60 distinct datasets derived from the rules extracted from DTs. These datasets were used to train the NN, alongside the final dataset, to assess how different combinations of rules and tree depths influence the NN's performance. \autoref{tab:dataset_summary} provides the details about the dataset used in the training process.

\begin{table*}[htbp]
    \centering
    \caption{Summary of Generated Dataset Configurations}
    \label{tab:dataset_summary}
    \begin{tabular}{|l|p{8cm}|c|c|}
        \hline
        \textbf{Dataset Type} & \textbf{Description} & \textbf{Count} & \textbf{DT Depths} \\ \hline
        Final Dataset & Baseline dataset used for evaluation & 1 & - \\ \hline
        Rule-Only Datasets & Extracted rules only & 12 & 3--15 \\ \hline
        Rules + Final Dataset & Rules and baseline dataset & 12 & 3--15 \\ \hline
        Rules (Variance 0.01) + Final & Rules filtered (variance $\leq$ 0.01) and combined with baseline & 12 & 3--15 \\ \hline
        Rules (Variance 0.001) + Final & Rules filtered (variance $\leq$ 0.001) and combined with baseline & 12 & 3--15 \\ \hline
        Rules (Variance 0.0001) + Final & Rules filtered (variance $\leq$ 0.0001) and combined with baseline & 12 & 3--15 \\ \hline
    \end{tabular}
\end{table*}

\section{Experiments and Results}
\label{sec:experiments}

\subsection{Evaluation Metrics}

In this study, we utilized three evaluation metrics to assess the performance of our model: R-squared ($R^2$), Mean Absolute Error (MAE), and Common Part of Commuters (CPC). 

\subsubsection{Mean Absolute Error (MAE)}
MAE is a metric that measures the average magnitude of errors in the model’s predictions, without considering their direction (i.e., positive or negative). It is given by the formula:

\begin{equation}
    MAE = \frac{1}{n} \sum_{i=1}^{n} |y_i - \hat{y}_i|
\end{equation}

where:
\begin{itemize}
    \item \(y_i\) is the actual value,
    \item \(\hat{y}_i\) is the predicted value,
    \item \(n\) is the number of data points.
\end{itemize}

\subsubsection{R-squared}

$R^2$, also known as the coefficient of determination, measures the proportion of variance in the dependent variable that is explained by the independent variables in the model. Mathematically, it is expressed as:

\begin{equation}
    R^2 = 1 - \frac{\sum_{i=1}^{n} (y_i - \hat{y}_i)^2}{\sum_{i=1}^{n} (y_i - \bar{y})^2}
\end{equation}

where:
\begin{itemize}
    \item \(y_i\) is the actual value,
    \item \(\hat{y}_i\) is the predicted value,
    \item \(\bar{y}\) is the mean of the actual values,
    \item \(n\) is the number of data points.
\end{itemize}

An R² value closer to 1 indicates that the model explains a higher proportion of the variance, implying better fit.

\subsubsection{Common Part of Commuters (CPC)}

The Common Part of Commuters (CPC), also known as the \textit{Sørensen-Dice index}, is a metric used to measure the similarity between the generated flows (\(y_g\)) and real flows (\(y_r\)) in flow generation models. CPC is computed as:

\begin{equation}
    CPC = \frac{2 \sum_{i,j} \min(y_g(l_i, l_j), y_r(l_i, l_j))}{\sum_{i,j} y_g(l_i, l_j) + \sum_{i,j} y_r(l_i, l_j)}
\end{equation}

where:
\begin{itemize}
    \item \(y_g(l_i, l_j)\) represents the generated flow from location \(l_i\) to location \(l_j\),
    \item \(y_r(l_i, l_j)\) represents the real flow from location \(l_i\) to location \(l_j\),
    \item The summations are over all possible origin-destination pairs \((i, j)\).
\end{itemize}

CPC ranges from 0 to 1, where 1 indicates a perfect match between the generated and real flows, and 0 signifies no overlap between them. When the total generated outflow equals the real total outflow, the CPC metric becomes equivalent to the accuracy of the model, measuring the fraction of trips assigned to the correct destination.

\subsection{Results}
\begin{figure*}[htbp]
    \centering
    \subfloat[Mean Absolute Error\label{fig:MAE}]{
        \includegraphics[width=0.65\columnwidth]{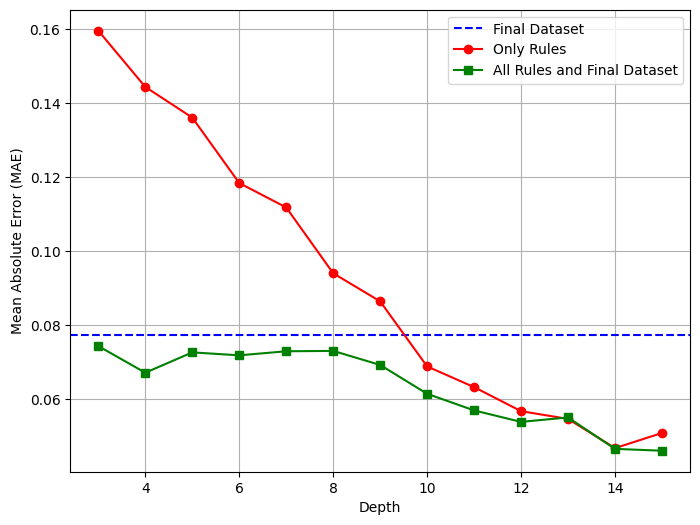}
    }
    \hfill
    \subfloat[$R^2$ \label{fig:R2}]{
        \includegraphics[width=0.65\columnwidth]{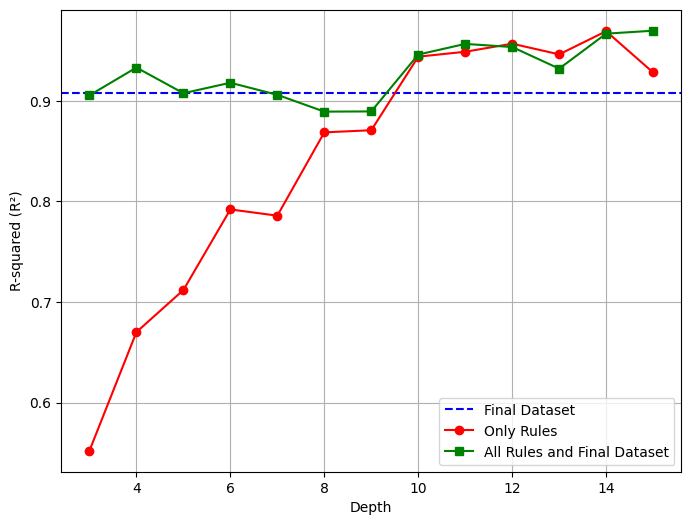}
    }
    \hfill
    \subfloat[Common Part of Commuters\label{fig:CPC}]{
        \includegraphics[width=0.65\columnwidth]{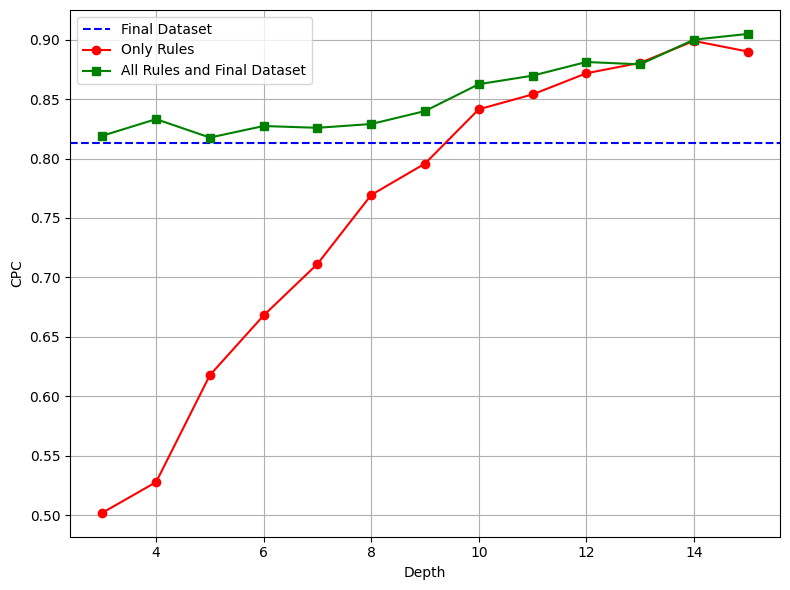}
    }
    \caption{NN performance on Final, Rules and Combined dataset}
    \label{fig:result1}
\end{figure*}

The \autoref{fig:result1} illustrates the performance of the NN model evaluated on three datasets: the Final Dataset, the Rules Dataset, and the Combined Dataset (Rules + Final Dataset). The results are analyzed using three metrics:  (MAE), \(R^2\), and CPC, across varying DT depths (3–15). The Figure\autoref{fig:MAE} demonstrates a significant reduction in MAE as the DT depth increases for the Rules Dataset and the Combined Dataset. The Rules Dataset initially exhibits high error, but deeper trees capture more granular patterns, reducing MAE. The Combined Dataset consistently achieves the lower MAE than the Final Dataset. However, the rules only dataset started to perform better consistently than baseline dataset in higher depths, notably after depth of 9. The \(R^2\) results of Figure\autoref{fig:R2} reveal the similar pattern. The Rules Dataset shows steady growth in \(R^2\), reflecting its ability to capture relationships more effectively as tree depth increases. The Combined Dataset achieves the highest \(R^2\) values across most depths, highlighting the advantage of integrating rules with the original dataset. The CPC results from Figure\autoref{fig:CPC} follow a similar trend, with the Rules Dataset and the Combined Dataset showing significant improvement as the tree depth increases. The Rules Dataset starts with lower CPC values, but as the rules grow more complex, it aligns better with the observed commuter patterns. The Combined Dataset achieves the highest CPC values at all depths, indicating superior alignment with commuting patterns.


\begin{figure*}[htbp]
    \centering
    \subfloat[Mean Absolute Error\label{fig:MAE_VarRules_Combned}]{
        \includegraphics[width=0.65\columnwidth]{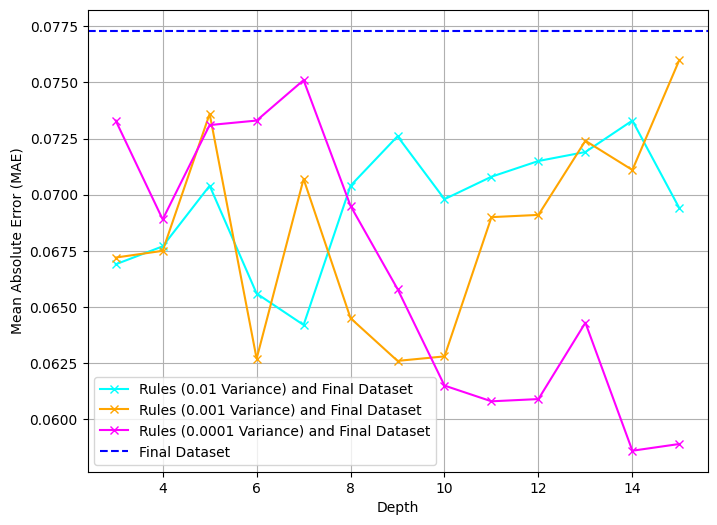}
    }
    \hfill
    \subfloat[$R^2$ \label{fig:R2_VarRules_Combned}]{
        \includegraphics[width=0.65\columnwidth]{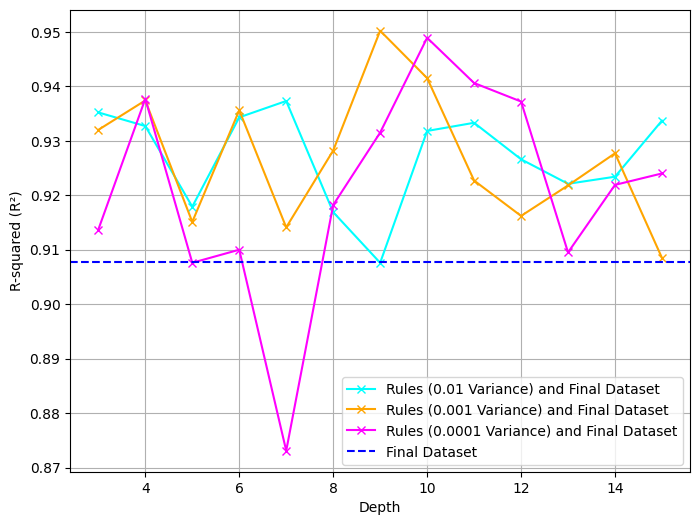}
    }
    \hfill
    \subfloat[Common Part of Commuters\label{fig:CPC_VarRules_Combned}]{
        \includegraphics[width=0.65\columnwidth]{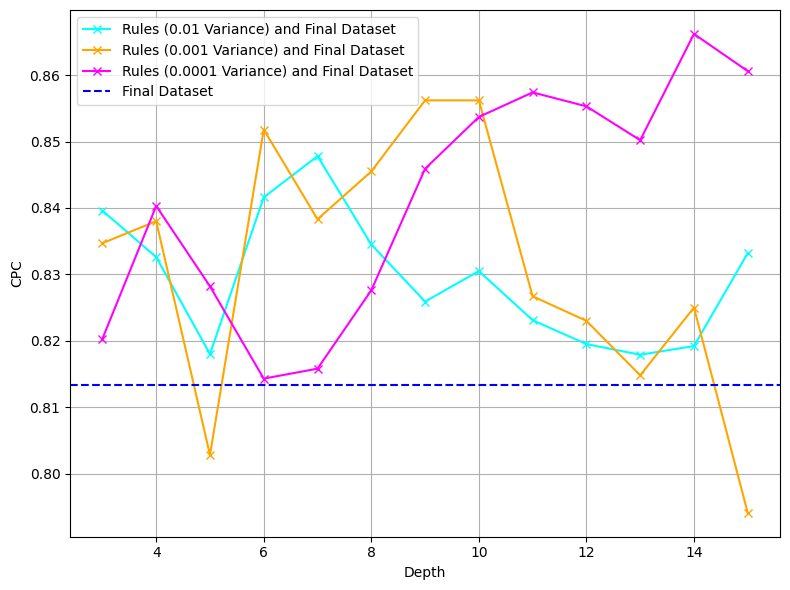}
    }
    \caption{NN performance on various sets of rules selected on the basis of variances}
    \label{fig:result2}
\end{figure*}

The  \autoref{fig:result2} illustrates the performance of the NN model evaluated on various sets of rules selected based on different variance thresholds (0.01, 0.001, 0.0001) in combination with the Final Dataset. The performance is compared using three metrics: MAE, \(R^2\), and CPC, across DT depths (3–15). The MAE in Figure\autoref{fig:MAE_VarRules_Combned} shows variability across different variance thresholds and depths. The rules selected with a variance threshold of 0.0001 consistently result in lower MAE values, indicating better predictive performance, particularly at greater depths. Conversely, rules selected with higher variance thresholds (0.01 and 0.001) exhibit higher MAE, especially at shallow depths. All the dataset are better than the baseline one. The \(R^2\) metric in Figure\autoref{fig:R2_VarRules_Combned} highlights the fluctuation in model performance based on different variance thresholds. The rules with a variance threshold of 0.0001 achieve the highest \(R^2\) values at most depths, indicating superior model fit. Rules with thresholds of 0.01 and 0.001 show fluctuating \(R^2\), reflecting inconsistencies in capturing relationships as depth increases. Similar pattern are seen in the Figure\autoref{fig:CPC_VarRules_Combned} that evaluates the alignment of predicted and observed commuter patterns. Rules with a variance threshold of 0.0001 exhibit consistently higher CPC values, particularly at greater depths, demonstrating their effectiveness in capturing commuter overlaps. Rules with thresholds of 0.01 and 0.001 show lower and more fluctuating CPC values, indicating reduced alignment. 

The  \autoref{tab:rules_table} provides an overview of the total rules generated, and the rules selected based on variance thresholds at various tree depths. The first two columns indicate the tree depth and the total number of rules generated (`All Rules`) at each depth. The subsequent columns represent the number of rules selected at different variance thresholds: 0.01, 0.001, and 0.0001.

As the tree depth increases, the total number of rules (`All Rules`) grows exponentially, reflecting the increasing complexity of the DT model. The rules selected by variance thresholds decrease with higher granularity (lower variance thresholds). For example, at a depth of 15, while 2628 rules are generated, only 5 rules are selected at the 0.01 variance threshold, 171 at 0.001, and 1595 at 0.0001. This trend underscores the importance of variance-based rule selection in simplifying the model while retaining the most relevant and impactful rules. The table shows the balance between tree depth and rule selection, with lower variance thresholds capturing more detailed patterns, offering a trade-off between complexity and interpretability.

\begin{table}[htbp]
\begin{center}
\caption{Rules and Variance-Based Rules at Different Tree Depths}
\label{tab:rules_table}
\begin{tabular}{|c|c|ccc|}
\hline
\multirow{2}{*}{\textbf{Depth}} & \multirow{2}{*}{\textbf{All Rules}} & \multicolumn{3}{c|}{\textbf{Rules Selected by Variance}} \\ \cline{3-5} 
                                &                & \textbf{0.01} & \textbf{0.001} & \textbf{0.0001}   \\ \hline
3                               & 8              & 6             & 8              & 8                \\ \hline
4                               & 16             & 7             & 14             & 16               \\ \hline
5                               & 32             & 11            & 24             & 32               \\ \hline
6                               & 64             & 15            & 32             & 58               \\ \hline
7                               & 116            & 18            & 46             & 101              \\ \hline
8                               & 202            & 20            & 60             & 163              \\ \hline
9                               & 324            & 20            & 83             & 251              \\ \hline
10                              & 505            & 23            & 103            & 374              \\ \hline
11                              & 753            & 20            & 128            & 531              \\ \hline
12                              & 1069           & 18            & 157            & 735              \\ \hline
13                              & 1485           & 14            & 164            & 992              \\ \hline
14                              & 2006           & 8             & 174            & 1278             \\ \hline
15                              & 2628           & 5             & 171            & 1595             \\ \hline
\end{tabular}
\end{center}
\end{table}

\begin{figure*}[htbp]
    \centering
    \subfloat[Mean Absolute Error\label{fig:MAEperrules}]{
        \includegraphics[width=0.65\columnwidth]{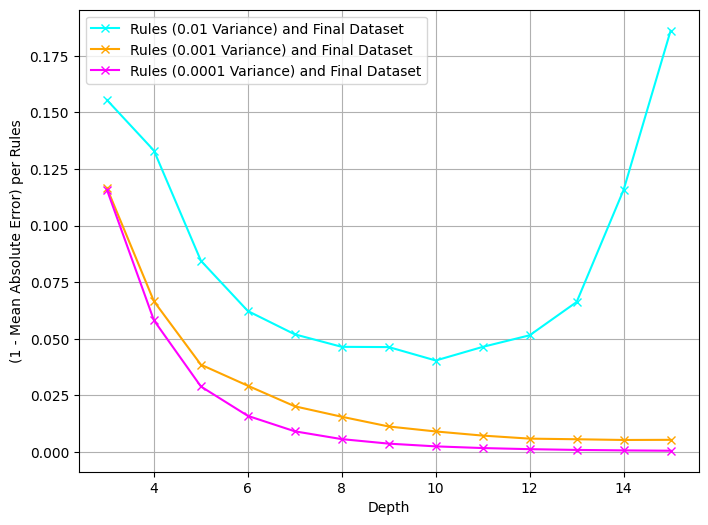}
    }
    \hfill
    \subfloat[$R^2$ \label{fig:R2perrules}]{
        \includegraphics[width=0.65\columnwidth]{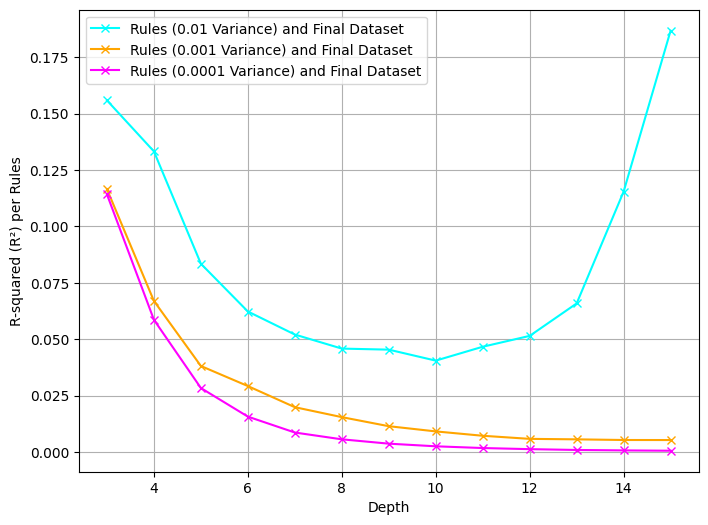}
    }
    \hfill
    \subfloat[Common Part of Commuters\label{fig:CPCperrules}]{
        \includegraphics[width=0.65\columnwidth]{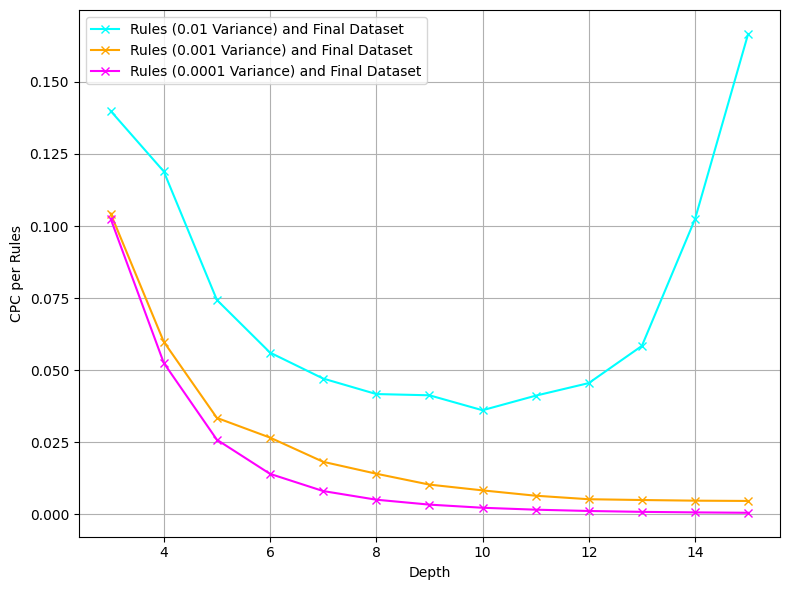}
    }
    \caption{NN performance per rules on different datasets}
    \label{fig:result3}
\end{figure*}

The  \autoref{fig:result3} illustrates the performance of the NN model per rule across different datasets, evaluated using three metrics: MAE, \(R^2\), and CPC, normalized by the number of rules. The datasets consist of rules selected based on variance thresholds (0.01, 0.001, 0.0001) combined with the Final Dataset. The analysis is conducted over tree depths ranging from 3 to 15. In Figure\autoref{fig:MAEperrules}, the performance per rule is higher in the lower depth as there are less number of rules which contributed highly. As the depth increased, more rules are generated, but the performance of the model doesn't increase in the same ratio, resulting in the less contribution per rule. This trend remains constant with variance 0.001 and 0.0001 as large number of rules are still being selected. But in the case of variance 0.01 though there are numerous rules generated but only few of them are selected as a result of which performance per rules increases after the depth of 10. In Figure\autoref{fig:R2perrules}, the \(R^2\) per rule shows a similar trend. Rules with a variance threshold of 0.0001 and 0.001 have better performance in lower depth.  Rules with thresholds of 0.01 show higher \(R^2\) per rule, in lower depth, which decreases with increase in depth and again rises after depth of 10. The Figure\autoref{fig:CPCperrules} depicts the almost identical pattern for  the CPC per rule. The lowest CPC per rule is observed at mid-range depths, with the performance increasing at both shallow and deeper depths for variance 0.01. Rules with a variance threshold of 0.0001 and 0.001 achieve the higher CPC per rule in lower depths, which keep on decreasing with the greater depths of DTs.


\section{Discussion}
\label{sec:discussion}
The experimental results demonstrate the effectiveness of integrating DT rules into the NN framework. Across all metrics— MAE, \(R^2\), and CPC—the combined dataset (Final Dataset + Rules) consistently outperformed the standalone datasets. This highlights the complementary nature of the original features and the extracted DT rules. Specifically, rules selected at lower variance thresholds (e.g., 0.0001) significantly enhanced performance, underscoring the importance of granular rule selection in capturing nuanced relationships within the data.

\subsection{Impact of Integrating Decision Tree Rules}
The integration of DT rules into the NN significantly improved its interpretability and predictive capability. The rules extracted from DTs provide explicit if-then conditions, enabling the model to incorporate domain knowledge and decision boundaries effectively. These rules not only complement the NN's ability to capture non-linear relationships but also mitigate overfitting by introducing structured and meaningful constraints. The experiments reveal that integrating rules led to lower MAE, higher \(R^2\), and improved CPC, particularly at greater tree depths and finer variance thresholds.

\subsection{Analysis of Rule Depth and Quantity}
The depth of the DTs and the number of rules extracted at each depth played a crucial role in the model's performance. As tree depth increased, the total number of rules grew exponentially, offering a richer representation of the data. However, this came with diminishing returns, as excessively large rule sets introduced noise and complexity. Rules selected at lower variance thresholds effectively balanced rule quantity and relevance, achieving superior performance. For instance, at depths 10–14, rules selected with a 0.0001 variance threshold consistently outperformed those selected with 0.01 or 0.001 thresholds, demonstrating that more granular rule sets capture finer patterns in the data.

\subsection{Potential Reasons for Observed Improvements}
Several factors contribute to the observed improvements in the model's performance:
\begin{itemize}
    \item Feature Augmentation: The integration of DT rules augmented the original feature set with interpretable and data-driven constraints, enhancing the model's ability to generalize.
    \item Granular Rule Selection: Rules selected at lower variance thresholds introduced finer-grained patterns, enabling the model to capture subtle relationships that might be overlooked in the original dataset.
    \item Hybrid Framework: Combining DT-based interpretability with the NN's capacity for learning non-linear relationships created a synergistic effect, leading to robust predictions.
    \item Reduced Overfitting: By incorporating rules, the model avoided over-reliance on noisy or irrelevant features, resulting in better generalization across depths.
\end{itemize}

\subsection{Limitations}
While the proposed methodology demonstrates promising results, certain limitations must be acknowledged to provide a balanced perspective and guide future improvements.

\subsubsection{Data-Related Constraints}
One key limitation lies in the potential biases or constraints inherent in the datasets used for this study. The reliance on data from specific sources, such as mobility datasets and economic indicators, may limit the generalizability of the findings to regions or scenarios with differing data availability or quality. Additionally, the static nature of the dataset may not fully capture temporal dynamics, such as seasonal or real-time fluctuations in travel demand.

\subsubsection{Rule Selection and Scalability}
The process of rule extraction and selection introduces challenges related to scalability and computational complexity. As tree depth increases, the number of generated rules grows exponentially, necessitating careful thresholding to ensure computational feasibility. However, this approach may inadvertently discard potentially useful rules, particularly at higher depths, leading to suboptimal feature representation.

\subsubsection{Interpretability Trade-offs}
Although DT rules enhance interpretability, integrating them with NNs may reduce the transparency of the overall hybrid framework. The black-box nature of NNs can obscure the relative contribution of individual rules, making it difficult to trace specific decision-making processes in certain scenarios.

\subsubsection{Methodology Bias}
The use of predefined variance thresholds for rule selection may introduce bias, as these thresholds are fixed rather than dynamically adapted to the data. This rigidity could lead to the exclusion of important rules or the inclusion of redundant ones, affecting the model's performance.

\section{Conclusion and Future Work}
\label{sec:conclusion}

This study proposed and evaluated a hybrid framework that integrates DT-based symbolic rules with NNs for travel demand prediction. The key findings demonstrate that incorporating DT rules significantly enhances model performance across multiple metrics, including MAE, \(R^2\), and CPC. Rules selected at finer variance thresholds (e.g., 0.0001) were particularly effective in capturing nuanced relationships, leading to improved predictive accuracy and alignment with observed commuter patterns. The integration of symbolic rules provided an interpretable structure while leveraging the non-linear learning capabilities of NNs, resulting in a synergistic performance boost. This approach aligns with the principles of Neurosymbolic AI, where symbolic rules enhance model transparency and provide actionable insights, while NNs enable the model to learn complex patterns from the data. The DT rules offer clear decision boundaries that complement the data-driven flexibility of NNs. This combination mitigates overfitting, augments the feature space, and improves generalizability.

Future research will aim to overcome the limitations identified in this study by exploring dynamic rule selection methods, adaptive variance thresholds, and more efficient mechanisms for integrating symbolic rules into NN architectures. Expanding the dataset to incorporate diverse, real-time data sources will enhance the framework's applicability across various scenarios. Furthermore, integrating explainable NN architectures, such as attention mechanisms or self-explaining models, could improve transparency and interpretability within the hybrid approach. 

\section*{Acknowledgment}

This material is based upon work supported by the NASA Aeronautics Research Mission Directorate (ARMD) University Leadership Initiative (ULI) under cooperative agreement number 80NSSC23M0059. This research was also partially supported by the U.S. National Science Foundation through Grant No. 2317117 and Grant No. 2309760.

\bibliographystyle{IEEEtran}
\bibliography{ref.bib}

\end{document}